\pdfoutput=1

\documentclass[11pt]{article}

\usepackage{booktabs}
\usepackage{graphicx,array}
\usepackage{multirow}
\usepackage{adjustbox}

\usepackage{acl}

\usepackage{times}
\usepackage{latexsym}

\usepackage[T1]{fontenc}

\usepackage[utf8]{inputenc}

\usepackage{microtype}

\usepackage{inconsolata}

\title{Distinguishing Translations by Human, NMT, and ChatGPT: A Linguistic and Statistical Study}

\author{
    Zhaokun Jiang$^\dagger$\quad Qianxi Lv$^\dagger$\quad Ziyin Zhang$^\ddagger$\thanks{\texttt{daenerystargaryen@sjtu.edu.cn}}\quad Lei Lei$^\diamond$ \\
    $^\dagger$ School of Foreign Languages, Shanghai Jiao Tong University\\
    $^\ddagger$ Department of Computer Science and Engineering, Shanghai Jiao Tong University\\
    $^\diamond$Institute of Corpus Studies and Applications, Shanghai International Studies University\\
}

\begin{document}
\maketitle
\begin{abstract}
The growing popularity of neural machine translation (NMT) and LLMs represented by ChatGPT underscores the need for a deeper understanding of their distinct characteristics and relationships. Such understanding is crucial for language professionals and researchers to make informed decisions and tactful use of these cutting-edge translation technology, but remains underexplored. This study aims to fill this gap by investigating three key questions: (1) the distinguishability of ChatGPT-generated translations from NMT and human translation (HT), (2) the linguistic characteristics of each translation type, and (3) the degree of resemblance between ChatGPT-produced translations and HT or NMT. To achieve these objectives, we employ statistical testing, machine learning algorithms, and multidimensional analysis (MDA) to analyze Spokesperson's Remarks and their translations. After extracting a wide range of linguistic features, supervised classifiers demonstrate high accuracy in distinguishing the three translation types, whereas unsupervised clustering techniques do not yield satisfactory results. Another major finding is that ChatGPT-produced translations exhibit greater similarity with NMT than HT in most MDA dimensions, which is further corroborated by distance computing and visualization. These novel insights shed light on the interrelationships among the three translation types and have implications for the future advancements of NMT and generative AI.
\end{abstract}

\section{Introduction}

\begin{figure}[th]
    \centering
    \includegraphics[width=1\linewidth]{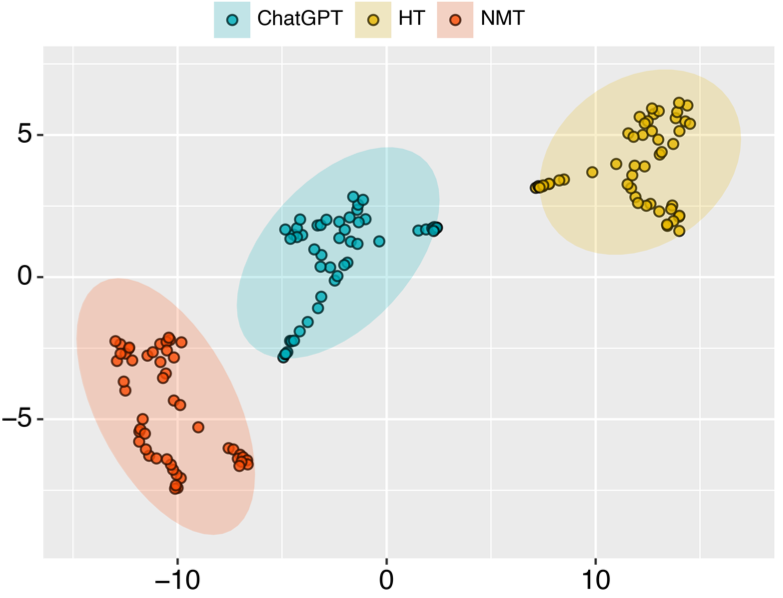}
    \caption{T-SNE visualization of distance distributions between translations by human, NMT, and ChatGPT.}
    \label{fig:fig10}
\end{figure}

Neural machine translation (hereafter NMT) has become an integral part of many translators and has garnered increasing attention from researchers in the field of translation studies~\citep{Gaspari2015}. Based on the powerful encoder-decoder architecture, NMT exhibits satisfying translation quality across a range of text types, and even showcases instances of creative translation~\citep{Hu-Li2023}. 

The advent of large language models (LLMs), exemplified by ChatGPT, has further revolutionized the dynamics between translation technology and translation practitioners. Existing literature has shown that ChatGPT can rival or even surpass the advanced NMT engines such as Google Translate and DeepL under both automated metrics and human evaluation, and is capable of producing natural, fluent, and human-like translations~\citep{Hendy2023}. As a promising AI-driven instrument and great contender to NMT, ChatGPT has attracted much attention from translation quality assessment (TQA) research that compares its performance against NMT using various metrics. There are also heated interests in improving its translation quality through prompt engineering~\citep{Wang2023,Peng2023,Lu2023,chen2023,He2023}.

Despite the promising role of ChatGPT as a powerful translation instrument, it is drastically different from NMT engines in various aspects. First, in terms of model architecture, it is a generative decoder-only model without the encoder part of NMT. In addition, while NMT is usually trained on parallel corpora mostly comprised of written language, the training data for ChatGPT are general-domain monolingual corpora in multiple languages containing a large proportion of web documents, dialogues, or even code. Also, ChatGPT is designed to be a versatile language model capable of performing various language-related tasks such as math reasoning~\citep{2021GSM8K} and coding~\citep{2023codesurvey}, whereas NMT is specifically used for translation. It thus deserves further inquiry how these differences may systematically influence the style, pattern, and property of their translation output.

According to our knowledge, limited attention has been given to in-depth analyses of linguistic and stylistic features that can differentiate ChatGPT-generated translations from NMT and human translation (hereafter HT). It also remains underexplored the relationship among these different types of translation, and whether ChatGPT-produced translation bears greater resemblance to HT or to NMT. Efforts to fill these gaps are important, because the widespread ``automation anxiety''~\citep{Vieira2018} and concerns about being replaced are to a large extent attributed to our lack of understanding of these advanced translation instruments. Built on the previous studies that compare NMT and HT from various perspectives, this study extends the investigative scope to incorporate ChatGPT, and explores their relative characteristics as exhibited by lexical and grammatical features on political translation.

Thus, we construct the major research questions of this study as follows:

(1) Can ChatGPT-generated translation be distinguished from NMT and HT? 

(2) What are the distinctive linguistic properties of NMT, ChatGPT-generated translations, and HT respectively?

(3) Do translations by ChatGPT bear more resemblance to HT or NMT?

To answer these questions, this research taps into the strength of corpus linguistics and builds a customized corpus comprising English translations of Spokesperson’s Remarks on important domestic and foreign affairs. We focus specifically on this register out of the following consideration: Spokesperson’s Remarks are unique in their combination of spontaneity and formality, since they are spontaneous answers to questions from reporters, but should be assertive and satisfying the requirements of the Chinese government at the same time. For this reason, typical features of spoken language and formal documents coexist in Spokesperson’s Remarks. It is this mixed characteristic that inspires us to explore how NMT, ChatGPT, and human translators will handle translation in this special register, and what are the distinctive differences among them as reflected in their language use.

We employ a multi-feature methodology by examining a group of relevant features at lexical, syntactic, and textual levels simultaneously. Specifically, we adopt widely applied machine learning methods, including classification and clustering techniques, together multidimensional analysis (MDA) proposed by \citet{Biber1988} to answer the first two research questions. Previous studies have already demonstrated the feasibility and necessity of analyzing translation through a simultaneous examination of multiple features~\citep{Hu2019,Kotze2016,Kotze2018,Calzada2022}. This approach also corresponds to a ``new, updated research agenda'' for translation studies~\citep{Sutter2019}, which calls for an interdisciplinary scope, a multimethodological framework, and an in-depth understanding of the multidimensional structure of translation~\citep{Calzada2022}. For the third question, we resort to the technique of distance calculation and dimension-reduced visualization to reveal the similarities among the three types of translation as measured by linguistic features. 

Our investigation provides tentative answers to whether ChatGPT can be an alternative translation tool apart from NMT, and demonstrates its own distinctive properties in comparison to NMT and HT. Such understanding may inform the future development of more human-like and contextually appropriate translation systems, and provide insights into when to rely on AI-generated translations, when to employ human translators, and when to combine both methods for more efficient and appropriate translations.
\section{Related Work}
\subsection{Machine Translation and Large Language Models}
Machine translation has a rich history in NLP, developing from statistical models~\citep{1990SMT,1993SMT} to neural models~\citep{2014seq2seq,2015attention} based on recurrent encoder-decoder architectures. \citet{2017Transformer} introduced Transformer with self-attention, which has now become the standard architecture in pretrained language models~\citep{2018GPT,2018BERT}, some of which scaled to hundreds of billions of parameters~\citep{2020GPT3,2022PaLM} and demonstrated early signs of artificial general intelligence~\citep{2023GPT4,2023AGI}.

While machine translation gave birth to the foundational architecture of large language models (LLMs), pretrained LLMs are also in return revolutionizing the learning paradigms in machine translation. \citet{2019GPT2}, for example, found that during the self-supervised pretraining of GPT-2, it implicitly learns to multitask (such as performing translation) from weak supervision signals in the pretraining corpus crawled from the web. While early pretrained language models such as T5~\citep{2019T5} are found to underperform dedicated NMT systems, the literature has recently witnessed much larger multilingual models such as BLOOM~\citep{2022BLOOM} and PaLM 2~\citep{2023PaLM2} that outperform supervised SOTA or even commercial translation engines, both by human evaluation and automatic metrics such as the traditional BLEU~\citep{2002BLEU} or model-based systems such as BLEURT~\citep{2020BLEURT}.

\subsection{Comparative Studies of NMT, ChatGPT, and HT}
The emergence of ChatGPT has introduced both new possibilities and challenges for NMT. Unlike traditional NMT systems that are constrained by the source language and its encoded representation, ChatGPT can generate translations in a more flexible manner, exhibiting more lexical diversity, syntactic variations, and textual adjustments. This can lead to fluent and context-aware but potentially less accurate translations, especially in cases where strict adherence to the source language is crucial~\citep{Hendy2023,Wang2023,Peng2023}. ChatGPT's decoder-only architecture and wide-ranging training data also make it more versatile. While NMT systems typically require parallel corpora for training, ChatGPT can be adapted to specific registers or languages using monolingual data alone, which makes it easily applicable for various translation tasks beyond traditional language pairs covered by parallel corpora~\citep{He2023,Kocmi2023}.

Numerous studies have already compared NMT and HT from various perspectives. Most of them focus on literary translation, aiming to identify the differences between the two translation approaches~\citep{Kuo2018,Frankenberg-Garcia2021,Hu-Li2023}. For instance, \citet{Kuo2018} examined the use of function words in machine-translated Chinese and in original Chinese, discovering an overuse of function words in MT. \citet{Frankenberg-Garcia2021} conducted a comparative lexical analysis of literary works translated by NMT and HT, revealing that HT to exhibited more explicitation, idiomaticity, register awareness, and risk aversion. In a comparison between Shakespearian plays translated by DeepL and human translators, \citet{Hu-Li2023} observed a certain degree of creativity in MT. More pertinent to our study is \citet{Sheng-Kong2023}, who examined the machine-translated Chinese political document in contrast to human translation, and found NMT to lack the subjectivity and flexibility of professional translators. These previous studies have provided valuable insights into the characteristics of NMT and HT. However, given the emergence and widespread adoption of ChatGPT, it is essential to expand the scope of comparison to include ChatGPT, so as to stay up-to-date with the rapid advancements in AI-powered language technology.

There are also abundant studies exploring the application of ChatGPT in translation tasks. One line of research focused on comparing the translation quality of advanced NMT engines and ChatGPT using automated metrics and human evaluation~\citep{Hendy2023,Raunak2023}. Their findings indicated that for high-resource language pairs, such as English and French, ChatGPT and GPT-4 could exhibit state-of-the-art translation capabilities, rivaling or outperforming the mainstream NMT systems. However, for low-resource language pairs and in highly domain-specific fields, ChatGPT still lagged behind NMT systems. \citet{Karpinska-Iyyer2023} showed that when translating paragraph-level texts, ChatGPT produced fewer mistranslations, grammatical errors, and stylistic inconsistencies compared to Google Translate. These results demonstrate the strong capabilities of ChatGPT in certain translation tasks, particularly for high-resource language pairs and general text translation. However, since ChatGPT is mainly trained on high-resource languages such as English, we are not fully aware of its competence in understanding and translating a middle-resource language like Chinese into English. Moreover, most of the previous assessments are conducted on publicly available corpora from OPUS or WMT, which makes other registers, such as diplomatic translation, underexplored fields awaiting more attention. 

In summation, while the existing literature has addressed the comparison among NMT, ChatGPT-generated translation, and HT from various perspectives, further investigation is still necessary to provide a more comprehensive understanding of their respective features. Such understanding is important since it allows us to optimize and leverage the capabilities of AI-powered language models in real-world translation scenarios, and facilitate the integration of human expertise and AI capabilities in the field of diplomatic translation.

\subsection{Multi-Feature Methods in Translation Studies}
Multi-feature analyses are commonly employed in corpus-based translation studies to explore the simultaneous effects of multiple relevant properties at lexical, syntactic, and textual levels. By considering multiple linguistic properties together, these methods provide a macroscopic view of various linguistic phenomena that cannot be captured by a single feature alone.

These methods are often applied in studies of translator attribution and translation stylistics in literary works. For instance, \citet{Rybicki-Heydel2013} attempted to attribute the correct translator in a collaborative translation that was completed by more than one translator. \citet{Mohamed2022} used machine learning algorithms to attribute Arabic translations of well-known literary books, aiming to identify which translator translated what texts. \citet{Wang-Li2011} compared two Chinese translations of Ulysses using parallel and comparable corpora. They analyzed keywords, lexical features, and syntactic features, concluding that translator fingerprints could be identified. Similarly, \citet{Fang-Liu2022} conducted a comparative study on three Chinese translations of Alice's Adventure in Wonderland. Their findings suggested that both translation style and translator style were visible and could be identified through multi-feature analyses.

Based on either customized or balanced corpora, multi-feature methodology also serves as a powerful tool for studies interested in distinguishing translational texts from non-translational ones. This line of research treats translated language as a distinctive type of language, often referred to as the ``third language''~\citep{Duff1981}, the ``third code''~\citep{Frawley1984}, ``constrained language''~\citep{Kotze2016}, or ``translationese''~\citep{Newmark1991}. A representative study was \citet{Hu2019}, who adopted a multi-feature statistical model adapted from MDA to examine differences between translated English and original English across registers. \citet{Kotze2016} used MDA to investigate the relationships between translated English and L2 varieties of English, and whether their shared features could be explained by constrained bilingual language production. Treating translation as a special language variety because of contact with other languages, \citet{Kotze2018} explored the influence of translation as well as other language varieties, register differences, and their combined effect in linguistic variation using MDA and a regression model. They identified register as the most significant factor in explaining linguistic variations. More related to our study is \citet{Calzada2022}, who only focused on one specific register, the parliamentary speech, to investigate linguistic variation by drawing on the identified dimensions in \citet{Biber1988}. 

These earlier studies have already delved into the characteristics of translated language from various perspectives, showing that multi-feature methods can be used to investigate linguistic variations across text types, registers and varieties of language. As neural machine translation and generative AI gain widespread adoption in recent years, scholars have paid more attention to identify their distinctive characteristics and patterns. Our study thus treats NMT and ChatGPT-generated translation as two different varieties of human translation, and aims to find their similarities and differences as exhibited in their respective translation output.
\section{Methodology}
The \textbf{methodology} of this study relies on a combination of computational techniques and multi-feature statistical analysis. Specifically, machine learning algorithms and Multidimensional Analysis (MDA, \citealp{Biber1988}) are employed to examine the relationships among three types of translation: ChatGPT-generated translations, NMT, and HT. The objective is to explore their shared characteristics, differences, and relative proximity to one another. Five major \textbf{procedures} are involved: (1) corpus building and text processing; (2) feature extraction; (3) the application of machine learning algorithms to classify the three types of translation; (4) the implementation of MDA and statistical models; (5) the calculation and visualization of distance among the three types of translation.

\subsection{Corpus Building and Text Processing}
The customized corpus in this study consists of three sub-corpora: (1) English translation made by institutional translators (Human\_Trans); (2) English machine translation by Google Translate (Machine\_Trans); (3) English translation generated by ChatGPT (ChatGPT\_trans). Their source texts are 147 pieces of spokesperson’s remarks, with each of them comprising questions proposed by foreign reporters and answers from the Chinese spokespersons centering several foreign affairs at a range of press conferences. 

We choose these textual materials out of three considerations: (1) data availability, as all the textual materials are readily accessible; (2) high-quality reference, since the human translation is performed by professional institutional translators working for the government and can serve as reliable reference; (3) complexity, because diplomatic discourse needs to demonstrate spontaneity and formality simultaneously.

All the textual materials can be directly downloaded from the official website of the Ministry of Foreign Affairs of the People’s Republic of China. Questions in these materials are asked in English, which are answered by the spokespersons in Chinese and then translated into other languages by institutional translators. Both the Chinese source texts and human translations underwent adjustment in wordings and contents, as well as corrections of speaking errors to be in line with the requirements of government websites. 

To build the sub-corpus NMT\_trans, the source texts are translated on the text level into English by Google Translate. The sub-corpus ChatGPT\_trans consists of English translations by ChatGPT\footnote{We use gpt-3.5-turbo-0613 API. Both NMT and ChatGPT translations are acquired on 20 October 2023.}. Basic information of the sub-corpora used in this study is listed in Table~\ref{tab:1}.

\begin{table}[th]
    \centering
    \begin{tabular}{lrr}
    \toprule
        Sub-corpus & Tokens & Samples\\
    \midrule
        Source & 59,505 & 147 \\
        Human\_Trans &38,697 & 147 \\
        Machine\_Trans &40,675 & 147 \\
        ChatGPT\_Trans &41,162 & 147 \\
    \bottomrule
    \end{tabular}
    \caption{Basic information of the sub-corpora.}
    \label{tab:1}
\end{table}

Since our objective is to uncover distinctive patterns inherent in the three types of translation, it is crucial to mitigate the external influence of stylistic differences attributed to spokespersons, text lengths, and contents. To be specific, as each spokesperson's remark may vary in length and topic, using their translations directly as text samples for frequency calculation could lead to misleading results. 

To address this concern, we resorted to a technique called rolling stylometry~\citep{Eder2016} to process the textual data. This approach involved the following steps: First, we concatenated all the translated texts in each sub-corpus into a single file. Next, we split the three concatenated files into equal-sized blocks of 5000 words. To ensure overlap and continuity between samples, we set the moving window size as 500 words. As a result, each sample, except the first and last ones, overlaps with its preceding and following samples. The remaining contents less than 5000 words were discarded. This process guaranteed that each sample is representative of its translation type, and is adequate for feature extraction. We demonstrate the operation of rolling stylometry in Figure 1.

\begin{figure}[t]
    \centering
    \includegraphics[width=1\linewidth]{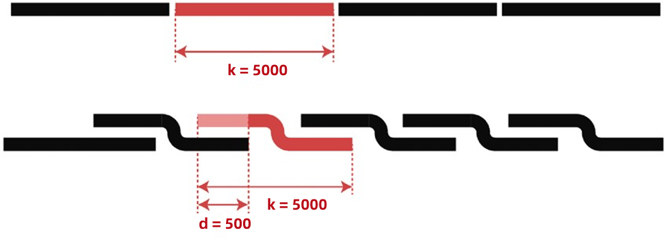}
    \caption{An illustration of rolling stylometry adapted from \citet{Eder2016}.}
    \label{fig:fig1}
\end{figure}

Following this procedure, we created a total of 210 samples: 67 from Human\_Trans, 71 from Machine\_Trans, and 72 from ChatGPT\_Trans. Then we conducted random sampling to select 50 samples for each sub-corpus. These samples are fundamental to this study, and will be repetitively used to address our three research questions.

\subsection{Feature Extraction}
Considering the uncertainty of which features may be significant to distinguish different types of translation, we follow \citet{Biber1988} and \citet{Hu2019} to incorporate as many standardized features as possible in the initial stage. We use MAT~\citep{Nini2019} and MFTE~\citep{LeFoll2023} to tag and extract 121 features in total. MFTE is developed as an extension of MAT to incorporate semantic tags from \citet{Biber1999} and \citet{Biber2006}. \citet{LeFoll2023} evaluated its performance in comparison to MAT and presented the reliability report\footnote{\url{https://github.com/elenlefoll/MultiFeatureTaggerEnglish/blob/main/Introducing_the_MFTE_v3.0.pdf}.}. The output is a csv file of normalized frequency counts. It will be used in the following machine learning experiments and MDA to address research question 1 and 2. The entire list of features and their descriptions can be found in the Appendix~\ref{sec:appendix-feature}. 

\subsection{Distinguishability Study via Clustering and Classification}
To examine whether the three types of translations are distinguishable, we employ several machine learning algorithms, both unsupervised and supervised. First, we apply unsupervised hierarchical clustering analysis (HCA) to group the text samples into clusters based on similarities in their feature patterns. The purpose is to examine to what extent can the text samples form three distinct clusters that correspond the three types of translation without training, and to unveil their potential separability.

A further investigation is undertaken by employing five supervised classifiers: Linear SVM, Random Forest, Multi-Layer Perceptron (MLP), AdaBoost, and Naïve Bayes, to perform a three-way classification task. We split our samples into a training set comprising 120 samples and a testing set containing 30 samples. If we see high accuracy, recall, and F1 scores, it means that the extracted features hold great discriminatory power to separate different types of translation. 

\subsection{Multidimensional Analysis}
To gain an in-depth understanding of the linguistic patterns of the three types of translation, we employ MDA~\citep{Biber1988} to analyze linguistic variation as reflected by the co-occurrence of extracted features. Based on the statistical technique of factor analysis, MDA is a ``bottom-up, data-driven'' method~\citep{Thompson2017} that considers registers, dimensions of co-occurring linguistic features, and text functions comprehensively. This study applies MDA to explore language variation among three different types of translation, namely, HT, NMT, and translations by ChatGPT. 

While the dimensions previously identified in \citet{Biber1988} could serve to compare HT, NMT and translations by ChatGPT, we decide to conduct a factor analysis from scratch, with an aim to identify context and register-specific factors in diplomatic translation. The procedure involves the following six steps: (1) selecting statistically significant linguistic features; (2) determining the number of factors based on scree plot; (3) performing factor extraction and factor rotation; (4) retrieving factor loadings; (5) interpreting the meaning of each factor; (6) comparing dimension scores of samples from HT\_trans, NMT\_trans, and ChatGPT\_trans. 

Not all the initially retrieved features are used in MDA, since we consider some features to be overlapping, and the multicollinearity issue may hinder the effectiveness of factor analysis. To avoid redundancy, we first examine whether or not each feature is statistically significant to distinguish the three types of translation. This is realized by conducting non-parametric Kruskal-Wallis H with a significance level of set at p < 0.05. To avoid Type 1 error caused by multiple comparisons, we use Bonferroni adjustment to adjust the p value. We also compute the pairwise correlations among the statistically significant features to exclude those exhibiting strong correlations with other features.

In addition, we carry out KMO Measure of Sampling Adequacy and Bartlett’s Test of Sphericity to examine the feasibility of our data for factor analysis. After that, we calculate eigenvalues, proportions of variance, and cumulative variance of each factor to determine the number of factors. 

To enhance the interpretability of factors, we perform the Varimax factor rotation. This method enforces orthogonality between factors, meaning that the factors are uncorrelated with each other. This simplifies the interpretation by ensuring that each factor represents a unique and independent dimension of the linguistic features. 

We also calculate the dimension scores of all the text samples within each factor by summing up the scores of positive features and subtracting those of negative features. We then standardize these scores using z-transformation, and use boxplots to display their relative positions along each dimension. 

\subsection{Calculation of the Pairwise Euclidean Distances and Visualization with t-SNE}
To investigate whether translations by ChatGPT were closer to HT or NMT, we calculate the pairwise Euclidean distance among these three types of translation using the z-transformed dimension scores, which results in three distance matrices. To represent the distance distribution, we use t-SNE (t-Distributed Stochastic Neighbor Embedding) for the purpose of visualization. T-SNE is a nonlinear dimensionality reduction technique that models the pairwise similarities between data points in a high-dimensional space, and maps them to a lower-dimensional space, where the similarities are preserved as much as possible. By using this technique, we can visualize the distances among the three types of translation in a reduced-dimensional space, and gain an intuitive understanding of the proximity of ChatGPT-generated translations to HT and NMT.

\section{Results and Findings}
\subsection{Clustering and Classification Results}
\begin{figure}[th]
    \centering
    \includegraphics[width=1\linewidth]{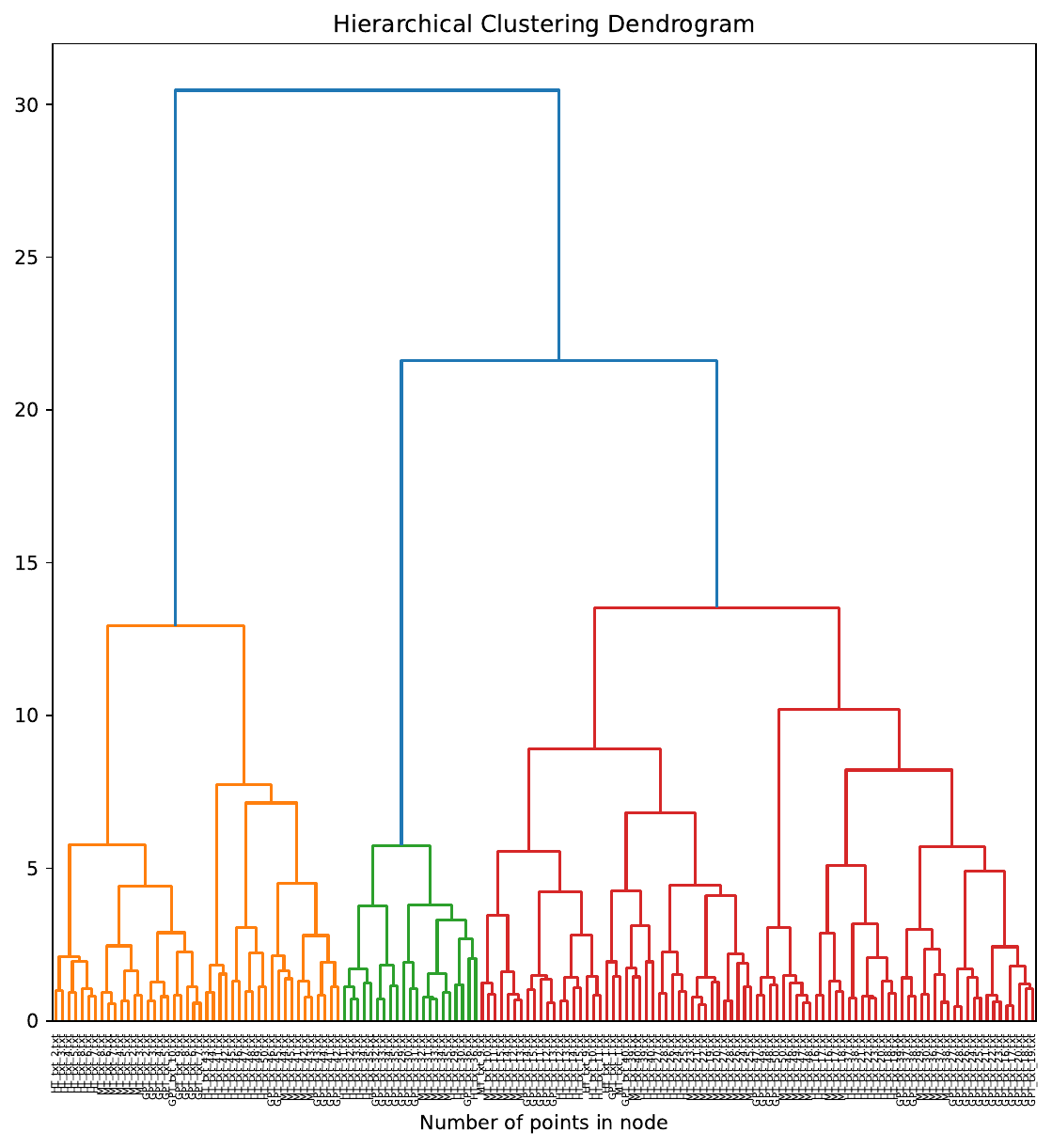}
    \caption{Hierarchical clustering dendrogram.}
    \label{fig:fig2}
\end{figure}

Figure~\ref{fig:fig2} shows that the text samples are roughly grouped into three distinct clusters. The largest cluster is situated on the right, whereas the smallest cluster is positioned in the middle. Zooming into the left cluster, we can see that the largest cluster is predominantly composed of NMT and ChatGPT samples, while the smallest cluster demonstrates a relatively balanced distribution of texts from HT, NMT, and ChatGPT. Overall, these results reveal that each cluster comprises samples from all three types of translation, suggesting that the unsupervised clustering technique seems unable to identify distinct patterns to differentiate the samples when labels are not attached to the translation samples. However, it does indicate that while HT, NMT, and ChatGPT-generated translations share much commonalities, they also differ from each other to some degree.

\begin{table*}[th]
    \centering
    \begin{tabular}{crcrrrr}
    \toprule
        Classifier & Accuracy & Class & Precision & Recall & F1 & Support \\
    \midrule
        \multirow{3}{*}{Linear SVM} & \multirow{3}{*}{0.97} & ChatGPT & 0.91 & 1.00 & 0.95 & 10 \\
        && HT & 1.00 & 1.00 & 1.00 & 9 \\
        && NMT & 1.00 & 0.91 & 0.95 & 11 \\
    \midrule
        \multirow{3}{*}{Random Forest} & \multirow{3}{*}{1.00} & ChatGPT & 1.00 & 1.00 & 1.00 & 10 \\
        && HT & 1.00 & 1.00 & 1.00 & 9 \\
        && NMT & 1.00 & 1.00 & 1.00 & 11 \\
    \midrule
        \multirow{3}{*}{MLP} & \multirow{3}{*}{1.00} & ChatGPT & 1.00 & 1.00 & 1.00 & 10 \\
        && HT & 1.00 & 1.00 & 1.00 & 9 \\
        && NMT & 1.00 & 1.00 & 1.00 & 11 \\
    \midrule
        \multirow{3}{*}{AdaBoost} & \multirow{3}{*}{0.97} & ChatGPT & 1.00 & 1.00 & 1.00 & 10 \\
        && HT & 1.00 & 0.89 & 0.94 & 9 \\
        && NMT & 0.92 & 1.00 & 0.96 & 11 \\
    \midrule
        \multirow{3}{*}{Naive Bayes} & \multirow{3}{*}{0.90} & ChatGPT & 0.77 & 1.00 & 0.87 & 10 \\
        && HT & 1.00 & 1.00 & 1.00 & 9 \\
        && NMT & 1.00 & 0.73 & 0.84 & 11 \\
    \bottomrule
    \end{tabular}
    \caption{Results of supervised classifiers with linguistic features.}
    \label{tab:tab2}
\end{table*}

Table~\ref{tab:tab2} presents the outcomes of the five classifiers on the test set. Overall, these results show the effectiveness of the classifiers in accurately classifying our samples into their respective classes. Notably, the Random Forest classifier and MLP classifier achieved full accuracy, precision, recall, and F1-scores for all classes, suggesting that HT, NMT, and ChatGPT samples are perfectly distinguishable. Linear SVM and AdaBoost gained full points on all the metrics for HT samples classification, but made a few mistakes on NMT and ChatGPT samples. Likewise, the Naïve Bayes classifier exhibited slightly lower scores for the NMT and ChatGPT class, but displayed perfect scores for the HT class, indicating the presence of a distinct boundary between HT and the other two types of translation. However, it also suggests that NMT and ChatGPT-produced translations may share some commonalities that cause the classifiers to misclassify them into the wrong categories.

\subsection{Interpreting Dimensions as a Result of Co-occurring Features}\label{sec4.2}
\begin{figure}[th]
    \centering
    \includegraphics[width=1\linewidth]{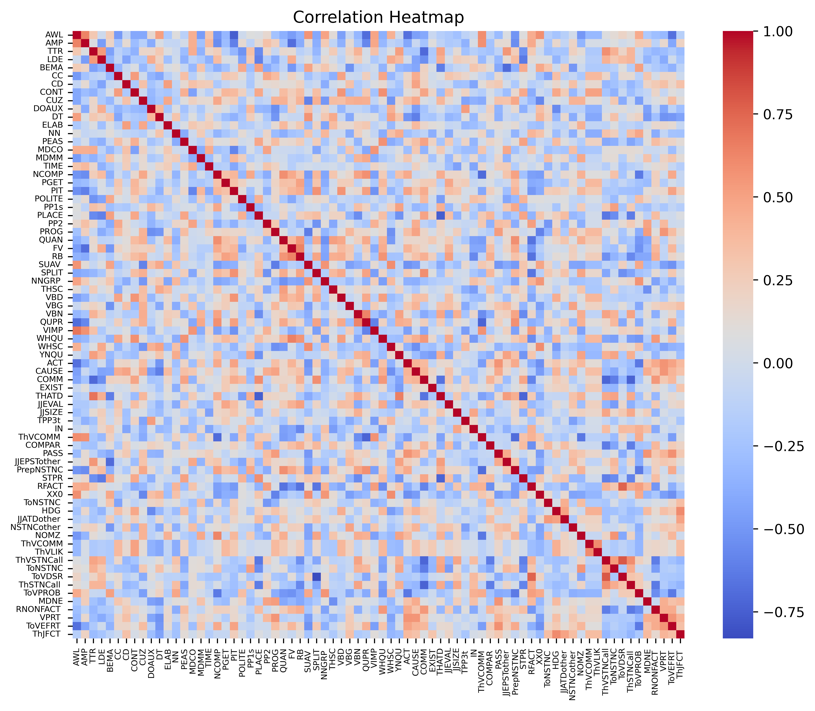}
    \caption{Correlation heatmap among the 74 statistically significant features.}
    \label{fig:fig3}
\end{figure}

Following the implementation of Kruskal-Wallis H, we found that 74 out of 121 linguistic features exhibited statistical significance The complete list of these features is in the Appendix~\ref{sec:appendix-significant}. As shown in Table~\ref{tab:tab3}, their frequency matrix demonstrates a KMO score of sampling adequacy exceeding the commonly employed threshold of 0.5 and thus ensures sampling adequacy. Additionally, Bartlett's test of sphericity yields a large chi-square value with a p value below 0.001. Therefore, we can safely proceed for the following analyses. However, as shown in Figure~\ref{fig:fig3}, ToVDSR and LDE are found to be highly correlated to other features, rendering the whole feature matrix a ``singular matrix''. In this case, factor analysis could not be conducted. To solve this issue, we made the decision to exclude these two features from the feature set, resulting in a remaining set of 72 features.

\begin{table}[]
    \centering
    \begin{tabular}{m{3cm}cr}
    \toprule
       KMO measure  & & 0.709 \\
       Bartlett’s test  & $\chi^2$ & 35663.45 \\
       & Significance & < 0.001 \\
    \bottomrule
    \end{tabular}
    \caption{Results of KMO and Bartlett’s test.}
    \label{tab:tab3}
\end{table}

\begin{figure}[th]
    \centering
    \includegraphics[width=1\linewidth]{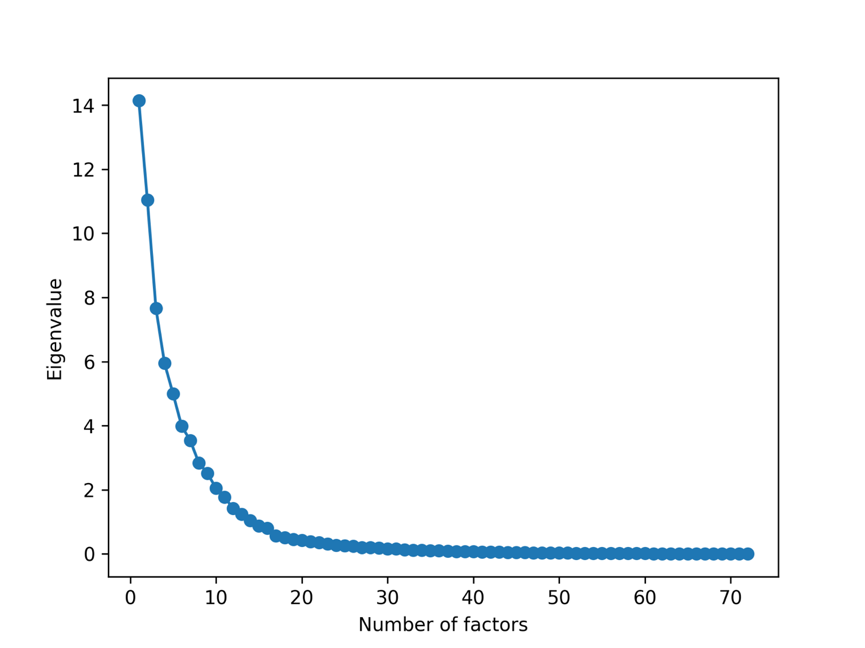}
    \caption{The scree plot showing the eigenvalue of the corresponding factor number. This helps determine the suitable number of factors to retain, as it identifies the point where the eigenvalues sharply decrease, indicating diminishing returns in terms of explained variance.}
    \label{fig:fig4}
\end{figure}

\begin{table*}[th]
    \centering
    \begin{tabular}{crrr}
    \toprule
        Factor & Eigenvalue & \% of variance explained & \% of cumulative variance explained \\
    \midrule
        0 & 14.0873 & 19.57 & 19.57 \\
        1 & 11.0070 & 15.29 & 34.85 \\
        2 & 7.6040 & 10.56 & 45.41 \\
        3 & 5.8799 & 08.17 & 53.58 \\
        4 & 4.9208 & 06.83 & 60.42 \\
        5 & 3.9331 & 06.46 & 66.88 \\
        6 & 3.4670 & 04.82 & 70.69 \\
        7 & 2.7698 & 03.85 & 74.54 \\
        8 & 2.4456 & 03.40 & 77.94 \\
        9 & 1.9909 & 02.77 & 80.70 \\
        10 & 1.6816 & 02.34 & 83.04 \\
    \bottomrule
    \end{tabular}
    \caption{Eigenvalue, proportion of variance and cumulative variance explained}
    \label{tab:tab4}
\end{table*}

The scree plot in Figure~\ref{fig:fig4} illustrates the eigenvalues associated with each factor, arranged in descending order. In Table~\ref{tab:tab4}, we present the corresponding eigenvalues, proportions of variance, and cumulative variance explained by each factor. Notably, the first five factors collectively account for almost 70 percent of the total variance. The primary factor alone explains approximately 20 percent of the variance, while the second factor contributes around 15 percent. From the sixth factor onwards, the explained variance decreases significantly, with less than 4 percent being accounted for. 

To determine the optimal number of factors, we manually evaluate the linguistic features within each dimension when considering a range of 4 to 10 factors. After careful examination, we conclude that the five-factor solution yields the most meaningful and interpretable outcomes. As a result, we have chosen to set the number of factors to five.

\begin{table*}[ht]
    \centering
    \adjustbox{width=\textwidth+1cm,center}{
    \begin{tabular}{lm{3.8cm}cm{10cm}}
    \toprule
        Dimension & Kruskal- Wallis H & \multicolumn{2}{l}{Features with positive/negative loadings} \\
    \midrule
        \multirow{4}{*}{1} & \multirow{4}{3.8cm}{$H(2, 150) = 31.43$, $p < 0.001$, $\eta^2= 0.605$} & positive & PASS (0.86), VBD (0.79), NN (0.76), NOMZ (0.71), AWL (0.71), NCOMP (0.69), WHSC 0.64), RFACT (0.62), TTR (0.55), IN (0.51), PEAS (0.47), THSC (0.41), DOAUX (0.34), POLITE (0.32), QUPR (0.31), PIT (0.30), SPLIT (0.30) \\
        \cline{3-4}
        && negative & VPRT (-0.79), PP1s (-0.77), ToNSTNC (-0.68), ELAB (-0.61), THAHD (-0.57), MDMM (-0.53), PP2 (0.49), PROG (-0.44), RB (-0.40), PRIV (-0.35), TIME (-0.33), PLACE (-0.31) \\
    \midrule
        \multirow{4}{*}{2} & \multirow{4}{3.8cm}{$H(2, 150) = 27.38$, $p < 0.001$, $\eta^2= 0.447$} & positive & ThSTNCall (0.87), COMM (0.85), THSC (0.81), ThVCOMM (0.74), MENTAL (0.71), THSC (0.70), MDWS (0.69), AMP (0.64), MDCO (0.59), WHQU (0.57), XX0 (0.47), PrepNSTNC (0.43), JJATDother (0.41), NSTNCother (0.39), HDG (0.36), CONC (0.31)\\
        \cline{3-4}
        && negative & ACT (-0.79), FV (-0.74), CAUSE (-0.63), ToVEFRT (-0.55), ToVDSR (-0.53), VBN (-0.46), CUZ (-0.43), RP (-0.38), CC (-0.32) \\
    \midrule
        \multirow{3}{*}{3} & \multirow{3}{3.8cm}{$H(2, 150) = 5.21$, $p = 0.07$} & positive &CD (0.73), CUZ (0.71), THATD (0.64), RFACT (0.57), ThJFCT (0.51), EXIST (0.44), NNGROUP (0.41), VBN (0.33), PROG (0.31), NNP (0.31), PGET (0.30)\\
        \cline{3-4}
        && negative &VLIKother (-0.70), ToVPROB (-0.48), RNONFACT (-0.42), ThJLIK (0.40), ELAB (-0.36), DT (-0.34) \\
     \midrule
        \multirow{2}{*}{4} & \multirow{2}{3.8cm}{$H(2, 150) = 21.67$, $p < 0.001$, $\eta^2= 0.425$} & positive &JJEVAL (0.64), JJEPSTother (0.55), VBG (0.51), THSC (0.46), JJSIZE (0.40), COMPAR (0.37), BEMA (0.37), TPP3t (0.31), ThVCOMM (0.30)\\
        \cline{3-4}
        && negative &STPR (-0.58), FV (-0.43), CAUSE (-0.36) \\
    \midrule
        \multirow{2}{*}{5} & \multirow{2}{3.8cm}{$H(2, 150) = 19.82$, $p < 0.001$, $\eta^2= 0.376$} & positive &SUAV (0.53), MDNE (0.49), CONT (0.42), NNGRP (0.33), VPRT (0.33)\\
        \cline{3-4}
        && negative & \\
    \bottomrule
    \end{tabular}
    }
    \caption{Features with positive and negative loadings in the five dimensions.}
    \label{tab:tab5}
\end{table*}

We only analyze features with loadings exceeding 0.30 or falling below -0.30. In Dimension 1, we have identified a total of 30 such features, as shown in Table~\ref{tab:tab5}. Notably, there exists a relatively larger cluster of features with positive loadings. The high frequency of nouns, noun compounds, nominalizations, and factive verbs indicates a high information density. High average word length and type/token ratio are often associated with rich vocabulary and elaborated language. Additionally, perfect aspect, past tense, passives, politeness markers, it pronoun reference, split auxiliaries, and infinitives can be indicators of a formal and polite tone.

For high-loading negative features, first person and second person pronouns are usually used in interactive discourse~\citep{Biber1988} and conversational scenarios. Modals may and might, to clauses preceded by stance nouns, together with private verbs serve as indicators of personal stance and attitude. Progressive aspect, time references, and place references are often used to denote concrete and specific events. Altogether, Dimension 1 can be interpreted as a dimension of precise information delivery with dense information and a formal tone at one end, and a rather interactive discourse in a relatively informal style at the other end.

\begin{figure}[th]
    \centering
    \includegraphics[width=1\linewidth]{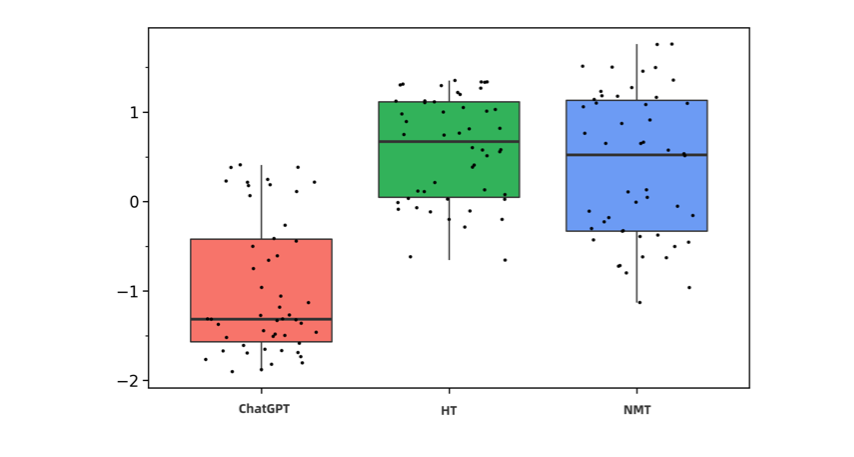}
    \caption{Z-transformed scores along Dimension 1.}
    \label{fig:fig5}
\end{figure}

The z-transformed scores of text samples in the three types of translation show that translations generated by ChatGPT contain more negative features in Dimension 1 compared with HT and NMT (Figure~\ref{fig:fig5}). This suggests that ChatGPT-generated translations are characterized by an engaged and interactive style with a relatively informal tone, whereas HT and NMT translations showcase a greater degree of formality and sophisticated language use. Their difference can be illustrated by the following examples:

(1) \textbf{It is reported that} Indian Prime Minister Narendra Modi visited the so-called "Arunachal Pradesh" on February 9th. […] \textbf{It has been verified that} there were eight Chinese citizens on board, including one from the Hong Kong SAR. (HT)

(2) \textbf{It is hoped that} all parties in Myanmar will proceed from the fundamental and long-term interests of the country and the nation, resolve emerging problems by peaceful means under the constitutional and legal framework, and continue to advance the process of democratic transition in the country in an orderly manner. (NMT) 

(3) \textbf{Really afraid of chaos in the world!} We can't help but ask these lawmakers, are you ``legislators'' or ``lawbreakers''? […] \textbf{You better mind your own business}. Hong Kong doesn't need you to \textbf{worry about} it. (ChatGPT)

Dimension 2 comprises a total of 25 features, consisting of 16 positive features and 9 negative features (Table~\ref{tab:tab5}). Noticeably, a wide range of positive-loading features in this factor are used to express stance (e.g., \emph{that} complement clauses preceded by a stance adjective or verb, \emph{will} and \emph{shall} modals, modal \emph{could}, amplifiers, negation, attitudinal adjectives, stance nouns, and hedges). Features such as direct WH-questions, communicative verbs often serve to engage with the others, delivering one’s own or asking for other people’s opinions. While amplifiers flag heightened emotions, hedges and concessive conjunctions can mitigate the intensity of attitudes. The most distinctive characteristic of these negative-weighted features is the prevalence of various types of verbs, including activity verbs, finite verbs, facilitation and causative verbs, and non-finite ed verb forms. Modals such as \emph{will}, \emph{shall}, and \emph{could} all indicate future action. Overall, this factor can be identified as one that distinguishes stance-oriented expressions from action-focused language.

\begin{figure}[th]
    \centering
    \includegraphics[width=1\linewidth]{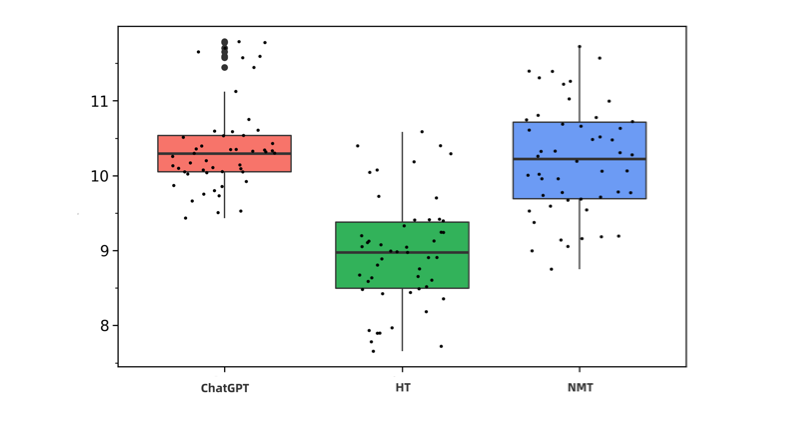}
    \caption{Z-transformed scores along Dimension 2.}
    \label{fig:fig6}
\end{figure}

Based on Figure~\ref{fig:fig6}, it is evident that translations generated by ChatGPT exhibit a considerable degree of similarity with NMT translations, while displaying significant differences from HT translations in Dimension 2. Human translators tend to employ fewer stance-related expressions and rely more on verbs, conjunctions, and coordinators. In contrast, translations produced by ChatGPT and NMT contain a higher frequency of linguistic features associated with the direct expression of stance and attitudes (e.g. negation, attitudinal adjectives, stance verbs, \emph{will} and \emph{shall} modals, and modal \emph{could}). 

Example (4), (5), and (6) below are provided to illustrate their distinctions. We can see that the human translator used three consecutive verb phrases to describe the efforts taken by the Algerian President to strengthen China-Algeria relations. In contrast, ChatGPT and NMT resorted to attitudinal adjectives ``appropriate,'' ``necessary,'' and ``important,'' attitudinal adverb ``resolutely,'' as well as the negation device ``no'' to convey a strong sense of stance taking. 

(4) During his term as Algerian President, he actively \textbf{promoted} the development of China-Algeria relations, \textbf{deepened} bilateral friendly cooperation and \textbf{enhanced} the friendship between the two peoples. (HT)

(5) China \textbf{will resolutely} take \textbf{appropriate} and \textbf{necessary} countermeasures according to the development of the situation. (ChatGPT)

(6) \textbf{No} force can stop the progress of the Chinese people and the Chinese nation. The most \textbf{important} criterion for judging \textbf{whether} the Chinese power situation is good or not is whether the Chinese people are \textbf{satisfied}. (NMT)

In Dimension 3, there are 12 features with positive loadings above the threshold of 0.3, including numbers, causal conjunctions, subordinator \emph{that} omission, factive adverbs, yes/no questions, nouns referring to group, \emph{that} subordinate clauses preceded by factive adjectives, existential or relationship verbs, nouns referring to human, non-finite \emph{ed} verb forms, progressive aspect, proper nouns, and get-passives. These features can be interpreted as devices to describe factual information or actual events that have already happened in an explicit and concrete manner.

The six features with negative loadings include likelihood verbs, to clauses preceded by verbs of probability, non factive adverbs, \emph{that} subordinate clauses preceded by likelihood adjectives, elaborating conjunctions, and determiners. The frequent co-occurrence of these features can be indicators of likelihood, possibility, and inference of future events. Taking both the positive and negative features into consideration, we can explain Factor 3 as distinguishing between factual description and inferential conjecture. 

\begin{figure}[th]
    \centering
    \includegraphics[width=1\linewidth]{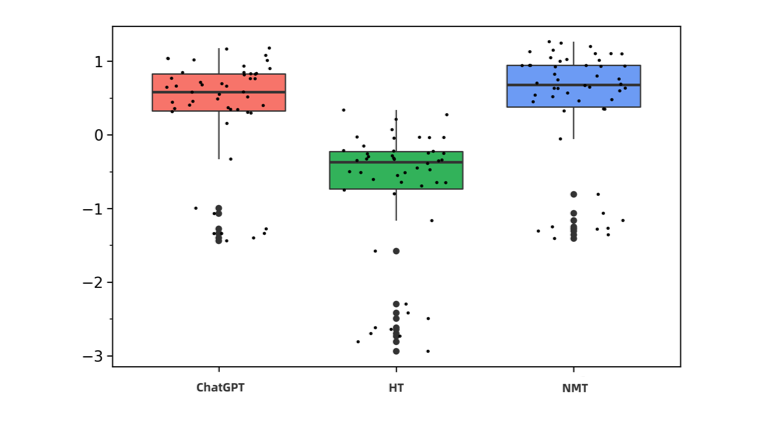}
    \caption{Z-transformed scores along Dimension 3.}
    \label{fig:fig7}
\end{figure}

From Figure~\ref{fig:fig7}, it is evident that both translations generated by ChatGPT and NMT lean towards the positive end of Dimension 3, indicating a slight preference for fact-oriented and information-focused language. In contrast, HT is more inclined to the negative end, suggesting that human translators use more expressions indicative of uncertainty and possibility. However, their differences do not amount to significance. 

From Example (7) and (8) below, we can see that translations by ChatGPT and NMT on the positive pole characterize frequent use of numbers, exact dates, and other factual information, while HT on the negative pole tends to convey likelihood (e.g. modal \emph{could}) rather than absolute facts, as shown in Example (9). 

(7) From 2010 to 2018, the Uyghur population in Xinjiang rose from \textbf{10,171,500} to \textbf{12,718,400}, an increase of \textbf{2,546,900}, an increase of \textbf{25.04\%}, which was not only higher than the \textbf{13.99\%} increase in the entire Xinjiang population, but also significantly higher than the \textbf{2\%} increase of the Han population. (NMT)

(8) According to reports, Tanzania’s National Electoral Commission released official results of the presidential election on \textbf{October 30} showing incumbent President John Pombe Joseph Magufuli winning another term with \textbf{84.3} percent of the vote. (ChatGPT)

(9) Should it choose to go further down the wrong path, it \textbf{could} expect more countermeasures from China. (HT)

Dimension 4 contains a total of 12 features, with 9 of them being positively weighted and 3 being negatively weighted. The positively loaded features include evaluative adjectives, epistemic adjectives without a that clause after, non-finite verb \emph{ing} forms, that subordinate clauses other than relatives, size-related adjectives, comparatives, \emph{be} as main verbs, reference to more than one non-interactant and single \emph{they} reference, and \emph{that} subordinate clauses preceded by communicative verbs. These features are associated with judgement and evaluative meanings, as well as comparison between entities or individuals

The three negatively weighted features consist of stranded prepositions, \emph{that} complement clauses not preceded by a stance adjective or verb, as well as auxiliary. They can be related to rather complicated sentence structure and non-evaluative discourse. We thus categorize this factor as a dimension characterizing a contrast between evaluative discourse and non-evaluative discourse. 

\begin{figure}[th]
    \centering
    \includegraphics[width=1\linewidth]{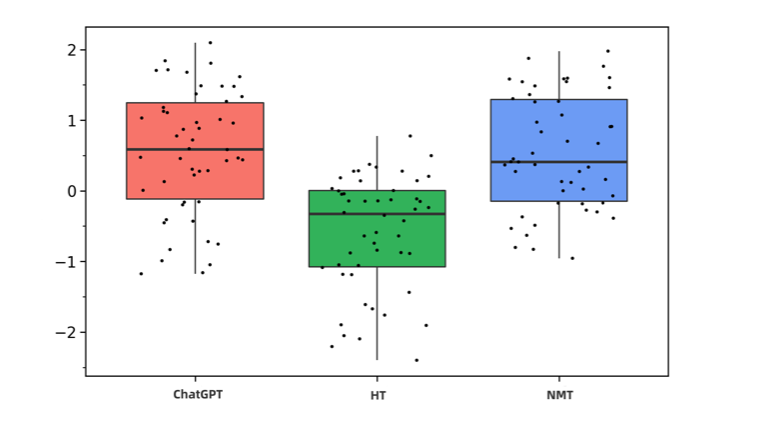}
    \caption{Z-transformed scores along Dimension 4.}
    \label{fig:fig8}
\end{figure}

Figure~\ref{fig:fig8} shows that ChatGPT-produced translations are more similar to NMT in Dimension 4. Both the two exhibit a higher frequency of positive features compared to HT translations. Their differences are clearly observable in Example (10), (11), and (12), which are translations from the same source text. We can see that ChatGPT uses the evaluative adjectives ``inappropriate,'' ``very bad,'' and ``unkind'' and \emph{be} as main verbs in its translation, to the effect that the evaluation is direct and intense. Similarly, NMT uses the epistemic adjective ``factual,'' and the evaluative adjectives including ``appropriate,'' ``very bad'' and ``unkind''. On the other hand, the human translator chooses different expressions such as ``neither in line with the facts nor out of place,'' ``go in the opposite way,'' and ``not a gesture of goodwill'' to convey the same meaning. These choices lead to a reduced intensity of evaluation compared to ChatGPT and NMT.

(10) In contrast, the words and actions of the US side not only do not conform to the facts, but are also \textbf{inappropriate}. The World Health Organization called on countries to avoid implementing travel restrictions, but before the words had even settled, the United States went against this and set a \textbf{very bad} precedent. That's really \textbf{unkind}. (ChatGPT)

(11) In contrast, the words and deeds of the US side are neither \textbf{factual} nor \textbf{appropriate}. The World Health Organization called on countries to avoid travel restrictions, but before the words fell, the United States did the opposite, with a \textbf{very bad} head. It's so \textbf{unkind}. (NMT)

(12) In sharp contrast, certain US officials' words and actions are \textbf{neither in line with the facts nor out of place}. Just as the WHO recommended against travel restrictions, the US rushed to go in the opposite way. Certainly \textbf{not a gesture of goodwill}. (HT)

Dimension 5 is the smallest in size, containing only five features with positive loadings. Among them, the most prominent one is suasive verbs, which is often used for the purpose of persuasion. Similarly, necessity modals convey the speaker’s strong personal conviction about a particular situation, and indicate a sense of obligation. They are often used to provide advice or make recommendations. Verbal contractions is typically seen in spoken discourse. Their co-occurrence alongside group nouns and present tense suggests that Factor 5 mainly revolves around the demonstration of strong will.

\begin{figure}[th]
    \centering
    \includegraphics[width=1\linewidth]{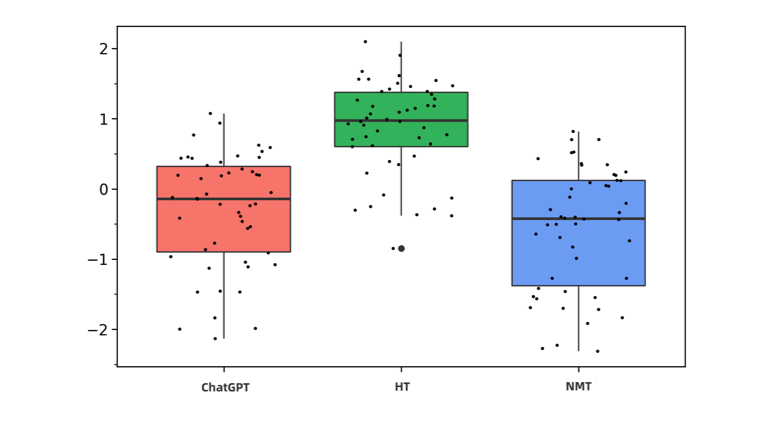}
    \caption{Z-transformed scores along Dimension 5.}
    \label{fig:fig9}
\end{figure}

Figure~\ref{fig:fig9} shows that HT scores the highest in this dimension, while NMT and ChatGPT-produced translations score lower. Also, the score distribution of HT is much more concentrated, suggesting that the dimensional characteristic is generally more noticeable in HT, which can be illustrated by the following example. 

(13) It \textbf{must} be pointed out that Hong Kong's prosperity and stability are in line with the interests of all parties, including the United States. […] The Chinese government and Chinese people are \textbf{firmly resolved} in safeguarding national sovereignty, security and development interests. (HT)

In summary, our factor analysis yields five meaningful dimensions. Four of them (Dimension 1, 2, 4, and 5) are able to distinguish ChatGPT-produced translations, NMT and HT with statistical significance. In Dimension 1 we observe a contrast between precise information in a formal style and interactive discourse with an informal tone. Dimension 2 in general differentiates expressions indicative of stance from language related to action taking. Dimension 4 characterizes a contrast between evaluative discourse and non-evaluative discourse, and Dimension 5 centers the demonstration of strong will. Measured by dimension scores, translations from ChatGPT seem to be more casual, explicit in stance-taking, and evaluative. NMT is mostly similar to ChatGPT-generated translations, but tends to be more formal, while HT features more formality and strong will but less evaluation. 

\subsection{Calculating and Visualizing Distances}
Based on the z-transformed dimension scores, the calculation of mean Euclidean distances between samples from ChatGPT and HT, ChatGPT and NMT, as well as HT and NMT are 4.233, 2.875, and 4.670 respectively. This implies that NMT is closest to ChatGPT-produced translations, and farthest from HT in a general sense. These findings are largely consistent with the results in Section~\ref{sec4.2}. The dimension-reduced t-SNE visualization is shown in Figure~\ref{fig:fig10}.
\section{Discussion and Conclusion}
Our first research question is whether translations by ChatGPT, NMT, and HT are distinguishable from each other. To answer this question, both unsupervised hierarchical clustering and supervised classification algorithms were employed. Notably, hierarchical clustering failed to separate text samples from different sub-corpora into distinct clusters. The implication is that the three types of translation do not have a clear-cut boundary, but are tangled with each other in most cases. However, with supervised training process, their distinctions and patterns became more discernable, though there were still cases where NMT and translations from ChatGPT were misclassified. We also found that HT was constantly classified correctly by all the supervised algorithms, suggesting that HT seems to be more easily identifiable compared with the other two types of translation. Based on these findings, we were confirmed that each type of translation possessed its own characteristics, but course-grained clustering or classification tasks were insufficient to unveil their exact distinctions. 

This thus led us to the second research question, which delved into the distinctive characteristics of ChatGPT-generated translations, NMT, and HT. We drew on MDA to conduct a fine-grained stylistic analysis by identifying and analyzing dimensions consisting of co-occurring features. We found that in four out of five dimensions, the z-transformed dimension scores of ChatGPT-produced translations were very close to those of NMT, as supported by their relative locations in the boxplots (see Figure~\ref{fig:fig6}, \ref{fig:fig7}, \ref{fig:fig8}, \ref{fig:fig9}) and the extracted text examples. In particular, we found that evaluative and attitudinal expressions were frequently observed in ChatGPT and NMT translations, but were less prevalent in HT. A possible reason is that human translators, who work for and represent the Chinese government, are proposed to be cautious in their language use, and adhere to the common practices in diplomatic translation. In other words, human translators are more risk-adverse~\citep{Pym2005} than NMT engines and ChatGPT. This is explainable, since both NMT engines and ChatGPT are in essence complicated neural networks, which resemble human brain but still lack the capacity to think as humans do. Therefore, their translations offer a semblance of “common sense”~\citep{Lee2023}, but lack the cultural sensitivity, linguistic flexibility, adaptability, and awareness of translation norms exhibited by human translators. As proposed by \citet{Pym2015}, an important technique in translators’ repertoire is “text tailoring,” where translators can change the content of the source text to better serve the purposes of the translated text. However, the current cutting-edge translation technologies, be it ChatGPT or NMT, are still unable to acquire such competency as making judgement and adaptation according to contexts and communicative needs.

A natural follow-up question addressed in this study is whether translations by ChatGPT are closer to NMT or HT. In line with our expectation, both the calculation of Euclidean distance and t-SNE visualization demonstrated that ChatGPT-generated translations were closer to MT, while HT was distant from both. The longest distance was observed between MT and HT. Similar observation was found in~\citep{Frankenberg-Garcia2021}, which offered a comprehensive analysis of the lexical differences between HT and NMT. The author found human translators to be superior in idiomaticity, the use of translation strategies, conveying register, and handling communication breakdowns. \citet{Karpinska-Iyyer2023} showed that paragraph-level translations by ChatGPT were more aligned with high-quality human translation, exhibiting reduced mistranslations, grammatical errors, and stylistic inconsistencies as compared to Google Translate. 

One implication of our analyses is that, even though NMT and generative intelligence represented by ChatGPT have made huge advances, there is still a marked gap between their translations and HT. There is thus a call for further investigations into the factors that contribute to the uniqueness and distinction of HT, encompassing not only formal linguistic properties but also other aspects. One possible way to improve the translation performance of ChatGPT and NMT is to identify the distinctive characteristics of top-notch translations, and then incorporate these characteristics into the training process of advanced AI-powered language models. Future research can focus on enhancing the adaptability and cultural awareness of automated translation tools, by creating translation technologies that combine the efficiency and speed of NMT and ChatGPT with the cultural sensitivity, linguistic flexibility, and domain expertise exhibited by professional human translators. This is crucial for the future development of NMT and artificial general intelligence (AGI) as a whole. Human translators, on the other hand, can make informed decisions when integrating these advanced translation technologies into their workflow. 

Lastly, we should acknowledge that this study is restricted in register and scope. Our investigation was conducted only in the field of Chinese-to-English diplomatic translation, and the relatively limited corpus size necessarily makes our analysis selective. Nevertheless, the findings may provide valuable insights into the characteristics of and relationship among ChatGPT-generated translations, HT, and MT. Moreover, the study also offers a set of methods and tools for future exploration with similar focuses.

\bibliography{custom}

\appendix

\section{Linguistic Features}\label{sec:appendix-feature}
In Table~\ref{tab:appendix-1} and \ref{tab:appendix-2}, we provide the 121 linguistic features used in this study.

\section{Statistical Significant Features}\label{sec:appendix-significant}
In Table~\ref{tab:appendix-3}, we record the 74 features that exhibit statistical significance.

\begin{table*}[th]
    \centering
    \adjustbox{width=\textwidth+1cm,center}{
    \begin{tabular}{m{2.5cm}m{14cm}}
    \toprule
        Category & Feature (Tag) \\
    \midrule
        General text properties & total number of words (Words), average word length (AWL), lexical diversity (TTR), lexical density (LDE), finite verbs (FV)\\
        \midrule
        Adjectives & Attributive adjectives (JJAT), predictive adjectives (JJPR)\\
        \midrule
        Adverbials & frequency references (FREQ), place references (PLACE), time references (TIME), other adverbs (RB) \\
        \midrule
        Determinatives & s-genitives (POS), determiners (DT), quantifiers (QUAN), numbers (CD), demonstrative pronouns and articles (DEMO) \\
        \midrule
        Discourse organizations & elaborating conjunctions (ELAB), coordinators (CC), causal conjunctions (CUZ), concessive conjunctions (CONC), conditional conjunctions (COND), discourse/pragmatic markers (DMA), filled pauses and interjections (FPUH), direct WH-questions (WHQU), question tags (QUTAG), yes/no question (YNQU), that relative clauses (TRHC), that subordinate clauses (other than relatives) (THSC), subordinator that omission (THATD), WH subordinate clauses (WHSC) \\
        \midrule
        Lexis & Total nouns (including proper names) (NN), noun compounds (NCOMP), hashtags (HST), superlatives (SUPER), comparatives (COMPAR), nominalization (NOMZ) \\
        \midrule
        Negation & Negation (XX0) \\
        \midrule
        Prepositions & Prepositions (IN)\\
        \midrule
        Pronouns & Reference to the speaker/ writer (PP1S), Reference to the speaker/ writer and others (PP1P), reference to addressee(s) (PP2), it pronoun reference (PIT), any personal pronoun not included in the other categories (PPOther), single, male third person reference (PP3m), single, female third person reference (PP3f), reference to more than one non-interactant and single they reference (TPP3t), quantifying pronouns (QUPR) \\
        \midrule
        Stance-taking devices & Politeness markers (POLITE), amplifiers (AMP), downtoners (DWNT), emphatics (EMPH), hedges (HDG)\\
        \midrule
        Stative forms & existential there (EX), be as main verb (BEMA)\\
        \midrule
        Verb features & Verbal contractions (CONT), particles (RP), be-passives (PASS), get-passives (PGET, going to constructions (GTO), past tense (VBD), non-finite verb-ing forms (VBG), non-finite ed verb forms (VBN), imperatives (VIMP), present tense (VPRT), perfect aspect (PEAS), progressive aspect (PROG), have got constructions (HGOT) \\
        \midrule
        Verb semantics & do auxiliary (DOAUX), necessity modals (MDNE), modal can (MDCA), modal could (MDCO), modals may and might (MDMM), will and shall modals (MDWS), modal would (MDWO), be able to (ABLE), activity verbs (ACT), aspectual verbs (ASPECT), suasive verbs (SUAV), facilitation and causative verbs (CAUSE), communication verbs (COMM), existential or relationship verbs (EXIST), mental verbs (MENTAL), private verbs (PRIV), public verbs (PUBV), Seem/appear (SMP), occurrence verbs (OCCUR), communicative verbs in other contexts (VCOMMother), factive verbs in other contexts (VFCTother), likelihood verbs in other contexts (VLIKother) \\
    \bottomrule
    \end{tabular}
    }
    \caption{Extracted linguistic features: part 1.}
    \label{tab:appendix-1}
\end{table*}

\begin{table*}[th]
    \centering
    \adjustbox{width=\textwidth+1cm,center}{
    \begin{tabular}{m{2.5cm}m{14cm}}
    \toprule
        Category & Feature (Tag) \\
    \midrule
        Adjectives semantics & attitudinal adjectives without a clause after (JJATDother), adjectives related to color (JJCOLR), epistemic adjectives without a that clause after (JJEPSTother), evaluative adjectives (JJEVAL), relational adjectives (JJREL), relational adjectives (JJREL), size related adjectives (JJSIZE), time related adjectives (JJTIME), topical adjectives (JJTOPIC) \\
        \midrule
        Adverb semantics & attitudinal adverbs (RATT), factive adverbs (RFACT), adverbs of likelihood (RLIKELY), non factive adverbs (RNONFACT) \\
        \midrule
        Noun semantics & Nouns abstracted and process (NNABSPROC), nouns cognitive (NNCOG), nouns concrete (NNCONC), nouns group (NNGRP), nouns human (NNHUMAN), nouns place (NNPLACE), nouns quantity (NNQUANT), nouns technical (NNTECH), nominalization (NOMZ), proper nouns (NNP), stance nouns without prepositions (NSTNCother) \\
        \midrule
        Syntax & that subordinate clauses (other than relatives) preceded by attitudinal adjectives (ThJATT), that subordinate clauses (other than relatives) preceded by adjectives of evaluation (ThJEVL), that subordinate clauses (other than relatives) preceded by likelihood adjectives (ThJLIK), that subordinate clauses (other than relatives) preceded by adjectives of evaluation (ThJEVL), that subordinate clauses (other than relatives) preceded by factive nouns (ThNATT), that subordinate clauses (other than relatives) preceded by factive nouns (ThNFCT), that subordinate clauses (other than relatives) preceded by attitudinal verbs (ThVATT), that subordinate clauses (other than relatives) preceded by communicative verbs (ThVCOM), that subordinate clauses (other than relatives) preceded by factive verbs (ThVFCT), that subordinate clauses (other than relatives) preceded by likelihood verbs (ThVLIK), mental/attitudinal verbs in other contexts (VATTother), to clauses preceded by ability adjectives (ToJABL), to clauses preceded by certainty adjectives (ToJCRTN), to clauses preceded by adjectives of ease (ToJEASE), to clauses preceded by factive adjectives (ToJFCT), to clauses preceded by evaluative adjectives (ToJEVAL), to clauses preceded by verbs of desire (ToVDSR), to clauses preceded by verbs of effort (ToVEFRT), to clauses preceded by mental verbs (ToVMNTL), to clauses preceded by verbs of probability (ToVPROB), to clauses preceded by verbs of speech (ToVSPCH), WH subordinate clauses preceded by attitudinal verbs (WhVATT), WH subordinate clauses preceded by communicative verbs (WhVCOM), WH subordinate clauses preceded by factive verbs (WhVFCT), WH subordinate clauses preceded by likelihood verbs (WhVLIK), To clauses preceded by stance nouns (ToNSTNC), prepositions preceded by stance nouns (PrepNSTNC), split auxiliaries and infinitives (SPLIT), stranded propositions (STPR) \\
    \bottomrule
    \end{tabular}
    }
    \caption{Extracted linguistic features: part 2.}
    \label{tab:appendix-2}
\end{table*}

\begin{table*}[th]
    \centering
    \begin{tabular}{lcclc}
    \toprule
        Feature & p value &\phantom{mmmmmmmm}& Feature & p value \\
    \midrule
        AWL & <0.001 & & WHQU & <0.001\\
    AMP & <0.001 & & WHSC & <0.001\\
    TTR & <0.001 & & YNQU & <0.001\\
    LDE & <0.001 & & ACT & <0.001\\
    BEMA & <0.001 & & CAUSE & <0.001\\
    CC & <0.001 & & COMM & <0.001\\
    CD & <0.001 & & EXIST & <0.001\\
    CONT & <0.001 & & THATD & 0.005\\
    CUZ & <0.001 & & JJEVAL & <0.001\\
    DOAUX & <0.001 & & JJSIZE & <0.001\\
    DT & 0.012 & & TPP3t  & <0.001\\
    ELAB & <0.001 & & IN & <0.001\\
    NN & <0.001 & & ThVCOMM  & <0.001\\
    PEAS & <0.001 & & COMPAR  & <0.001\\
    MDCO & <0.001 & & PASS & <0.001\\
    MDMM & <0.001 & & JJEPSTother  & <0.001\\
    TIME & <0.001 & & PrepNSTNC & <0.001\\
    NCOMP & <0.001 & & STPR & <0.001\\
    PGET & <0.001 & & RFACT & 0.013\\
    PIT & <0.001 & & XX0 & <0.001\\
    POLITE & <0.001 & & ToNSTNC  & <0.001\\
    PP1s & <0.001 & & HDG  & <0.001\\
    PLACE & <0.001 & & JJATDother  & <0.001\\
    PP2 & <0.001 & & NSTNCother  & <0.001\\
    PROG & 0.009 & & NOMZ  & <0.001\\
    QUAN & <0.001 & & ThVCOMM & <0.001\\
    FV & <0.001 & & ThVLIK & <0.001\\
    RB & <0.001 & & ThVSTNCall & <0.001\\
    SUAV  & <0.001 & & ToNSTNC & <0.001\\
    SPLIT & <0.001 & & ToVDSR & <0.001\\
    NNGRP  & <0.001 & & ThSTNCall  & 0.011\\
    THSC & <0.001 & & ToVPROB  & <0.001\\
    VBD & <0.001 & & MDNE & <0.001\\
    VBG & <0.001 & & RNONFACT  & <0.001\\
    VBN & <0.001 & & VPRT  & <0.001\\
    QUPR & 0.016 & & ToVEFRT  & <0.001\\
    VIMP & <0.001 & & ThJFCT  & <0.001\\
     
    \bottomrule
    \end{tabular}
    \caption{74 linguistic features with statistical significance.}
    \label{tab:appendix-3}
\end{table*}

\end{document}